# A User-Centred Framework for Explainable Artificial Intelligence in Human-Robot Interaction

**Marco Matarese,** [1,2] **Francesco Rea,** [2] **Alessandra Sciutti**[2]

[1]University of Genoa
[2]Italian Institute of Technology
Via Enrico Melen, 83 - Genova (GE) 16152
{marco.matarese, francesco.rea, alessandra.sciutti}@iit.it

## Abstract

State of the art Artificial Intelligence (AI) techniques have reached an impressive complexity. Consequently, researchers are discovering more and more methods to use them in real-world applications. However, the complexity of such systems requires the introduction of methods that make those transparent to the human user. The AI community is trying to overcome the problem by introducing the Explainable AI (XAI) field, which is tentative to make AI algorithms less opaque. However, in recent years, it became clearer that XAI is much more than a computer science problem: since it is about communication, XAI is also a Human-Agent Interaction problem. Moreover, AI came out of the laboratories to be used in real life. This implies the need for XAI solutions tailored to non-expert users. Hence, we propose a user-centred framework for XAI that focuses on its social-interactive aspect taking inspiration from cognitive and social sciences' theories and findings. The framework aims to provide a structure for interactive XAI solutions thought for non-expert users.

## Introduction

Artificial Intelligence (AI) techniques - especially Machine Learning (ML) ones - are being increasingly used in several areas of our society (Kotsiantis et al. 2006). This massive use of ML-based systems in turn underlined the need to make such systems transparent to the human user, which means making clear the rationale behind their decisions to those who use them. This is key especially when ML systems are used in crucial areas, such as insurance or credit application (Gunning and Aha 2019).

To this need, which is becoming more and more urgent as the use of ML systems increases, the AI community has responded with the introduction of a new research field: the Explainable AI (XAI) (Vilone and Longo 2020). Among the enormous number of different ML techniques, those based on neural networks have been considered as more opaque than other techniques (Adadi and Berrada 2018). Thus, most of the efforts in XAI have been directed to make neural networks explainable (Montavon et al. 2018).

However, what an explanation is and what form it should take is not a trivial matter. The first attempts used explanations that sounded reasonable to the developers, without any kind of scientific support about the effectiveness of such explanations for non-expert users (Miller et al. 2017). Therefore, Miller (2019) proposed to bring insights from social sciences to build tailored-to-humans explanations. In his work, he pointed out that XAI is a Human-Agent (-Robot) Interaction (HRI) problem as well as a computer science one. Indeed, to explain is a social problem because it assumes the form of communication between individuals. As such, explanations have to comply with the most common social rules (Hilton 1990). Hence the need to take from social sciences insights about both form and content of explanations, as well as how to provide them.

People explain themselves to other people every day: robots should exploit these habits in XAI scenarios. On the other hand, people attribute human-like traits to artificial agents, thus they expect from them explanations using the same conceptual framework they are used to (De Graaf and Malle 2017). To develop those systems taking in mind humans' needs means to make XAI human-centred (Hellström and Bensch 2018).

Most of the solutions proposed so far have not reached a complete human-centred approach (Miller 2019). Moreover, even if the number of works following this philosophy is rapidly increasing, a more generalist approach is missing. Indeed, the most complex XAI solutions are designed to be ad-hoc in terms of the AI technique they seek to render explainable, the context in which the system is used, and who they are aimed at.

In this work, we present a theoretical framework for XAI in HRI. Differently from other works in XAI (Ciatto et al. 2020; Sanneman and Shah 2020), our framework models the explanation process as an interaction between the two parties. In addition, its explanations generation uses social and cognitive science's theories and recent findings of what people expect from explaining agents. Therefore, the framework aims to be user-centred, which means that it generates personalised explanations (Sokol and Flach 2020), but also contextualised once concerning users' needs. In particular, it considers users actions' effects, their information needs, and it uses human-inspired heuristics. We focused on *post-hoc* explanations (those asked after the robot action) because they are the most common type of explanations in people's everyday lives and the more interesting one from the social-interactive perspective (Lipton 2016).

## Background and Related Works

Lewis (1986) defined the act to explain an event as *"to provide information about its causal history"*: someone who has such information tries to convey it to someone else. Moreover, causality and counterfactuals - which are states of the world that would have resulted from events that did not occur - are notions related to explanations. Hume (2000), in his regularity theory of causation, claims that there is a cause between two types of events if events of the first type are always followed by events of the second. However, Hume's definition is about counterfactuals rather than just dependence. He argued that the occurrence of two events $A$ and $B$ do not give useful causal information. The cause should be understood relative to a counterfactual case: the event $A$ causes $B$ if, under a hypothetical counterfactual case, the event $A$ did not occur, the event $B$ would not have occurred (Hume et al. 2000). Several causality models have been proposed in the years that focus on counterfactuals (Halpern and Pearl 2005; Menzies and Price 1993; Fair 1979). Furthermore, the concept of causal chains is an important component in explanation. Causal chains are defined as paths of causes that bring together a set of events (Hilton et al. 2005). However, people do not need to understand the complete causal chain of an event to have a good explanation.

Lombrozo (2006) argued that explanation is both a process and a product. It is a *cognitive* process because of the abductive inference for filling the gaps to determine the explanation of an event (Chin-Parker and Bradner 2010), but also a *social* process because of the transfer of knowledge between who asks for explanations - the *explainer* - and who provides them - the *explainee* (Miller 2019). Moreover, an explanation is the product of the cognitive process. In particular, explanations are answers to *why-questions* (Dennett 1989; Lewis 1986). Why-questions are *constrastive* (Lipton 1990) meaning that the *explainee* does not want to be explained the occurrence of the event *per se*, but why it occurred in that case and not in some other counterfactual cases (Hilton 1990). Several pieces of research confirmed the hypothesis by which people tend to ask questions about events that they consider unexpected (Heider 2013). For this reason, the *explainer* must understand the counterfactual case.

Early attempts at XAI lacked effectiveness for the end-users mainly because they were designed and intended for computer scientists. This led to solutions that developers intuitively thought were useful for them, without any objective measure of effectiveness. Nevertheless, in recent years, researchers began to focus on the design and evaluation of good explanations (Mohseni et al. 2018). Furthermore, several works have been proposed trying to address the XAI problem in a new fashion. For instance, Chakraborti et al. (2017) proposed a way to align user and robot's models during planning tasks through models reconciliation. Instead, Tabrez et al. (2019) emphasised users' task understanding: ad-hoc explanations have the objective of correcting users' task understanding to detect incorrect beliefs about the robot's internal functioning. Moreover, Han et al. (2021) focused on organising and representing complex tasks in an interpretable way using behaviour trees.

Similarly, the number of abstract frameworks for XAI that consider users is rapidly increasing in the literature (Ciatto et al. 2020). However, even if much effort is being made to made XAI tailored-to-humans, a complete human-centred characterisation has not been reached so far (Miller 2019). Moreover, even if the number of studies following the social sciences inspired approaches is increasing (Ciatto et al. 2020), a more generalist approach is missing because researchers prefer to focus on ad-hoc solutions (Sanneman and Shah 2020). Recent works moved in the direction of an alignment between *explainer* and *explainee* to build a common ground on which base systems' explanations (Thellman and Ziemke 2021), while others focus on a wide spectrum of explainability social aspects (Ehsan et al. 2021; Wallkötter et al. 2021; Arnold, Kasenberg, and Scheutz 2021).

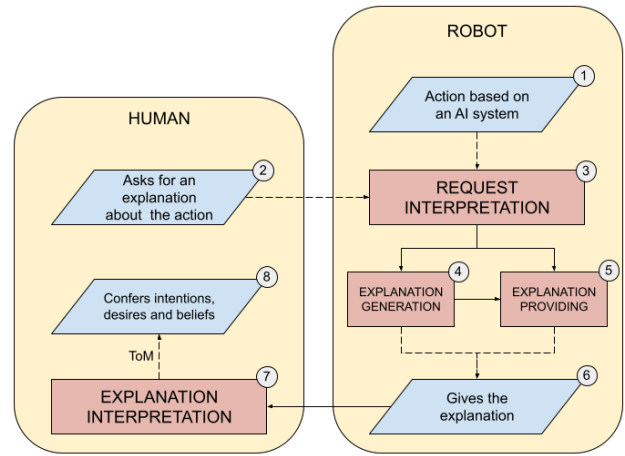

Figure 1: The framework. The boxes with trapezoidal shapes refer to agents' actions, while the rectangular ones are the framework's modules. The numbered labels indicate the temporal order of both actions and modules.

The framework we present in this paper has a twofold objective: modelling the interaction between the parties, and including cognitive science's theories and findings into XAI to generate users-centred explanations. This way, we aim to emphasise the social-interactive aspect of this discipline. Resting on solid foundations, given by social sciences, we aim to base robots' explanations on people's everyday habits and expectations in explanation interactions.

## The Framework

The awareness that the social-interactive aspect is key in XAI lead to interesting consequences. First of all, at least two agents are involved: the *explainer* and the *explainee*. In our case, the human takes the role of *explainee* and the robot the one of *explainer*. Figure 1 shows a graphical representation of the framework we propose. The boxes with trapezoidal shapes refer to agents' actions while the rectangular ones represent the framework's modules that allow both agents to perform the actions mentioned above. The numbered labels indicate the timing in which the framework's parts are involved. As mentioned, we focused on *post-hoc* explanations. The main characteristic of such explanations is that the *explainee* asks for explanations as soon as the *explainer* acted. For this reason, the request for explanations immediately follows the robot action. After the robot received the request, it should interpret such a request, choose an explanation to give, and communicate it. These three actions are managed by three different modules working in cascade: the *Interpreter*, the *Explanation Generation* and the *Explanation Providing* modules. Our framework should be thought of as supporting the AI system managing the robot's behaviour and that we want to make explainable.

### The Interpreter

The first thing an XAI system should do is to interpret the request for explanations. As mentioned above, such requests have always the form of a *why-question*. In particular, they are *contrastive* questions because something that the *explainee* did not expect has occurred. Hence, the interpreter receives the question as input and produces two outputs: the *fact*, which is the event that occurred, and the *foil*, a not occurred event that the human expected to happen. Due to its contrastive nature, it is always possible to produce a fact and a foil from a request for explanations. As a result, the interpreter transforms the request into the question "why *fact* rather than *foil*?". The generation of a fact and a foil could be particularly difficult when the *explainee* leaves the foil implicit. To handle these kinds of scenarios, the interpretation module should provide a mechanism to infer the implicit foil. Some authors proposed to simplify the problem by defining the foil as the negation of the fact: "why *fact* rather than not-*fact*?" (Lipton 1990). Other authors, claimed that we can use the concept of *normality* to address this problem: "why *fact* rather than the normal case *foil*?" (Hilton 1990).

### The Explanation Generation Module

Once the request has been interpreted, we need to produce an explanation that answers the interpreted question and sounds reasonable to the *explainee*. Hence, the Explanation Generation module is the core of the framework: it aims to find the most appropriate explanation given the fact and foil, which are both given as input to this module.

Figure 2 shows the details of the explanation generation. It consists of three sub-modules working in cascade. The first one aims to find the causal relationships between fact and foil and produces a set of causal chains. For this purpose, it uses also the objectives of the robot because they could have an essential role in explaining the robot's action selection. The second one aims to find the most appropriate causal chains given the context. During this step, the framework uses the users' information needs, and the state of both the robot and environment to set up a context. Finally, filtering is applied to the resulting explanations to select only one of them.

**Finding Causality** The first step of the module aims to understand what exactly the request for explanations refers to. As such requests could also be related to the objectives of the robot, this sub-module uses also those to relate fact and foil. In fact, the objectives of the robot have a crucial role in its behaviour. Thus, it could help in relating the fact and foil in a way that is easily understood by the end-user. This sub-module aims to produce causality chains: paths of causes between a set of events. Thus, the output of this sub-module consists of a set of causal relationships that could potentially clarify the rationale behind the robot actions.

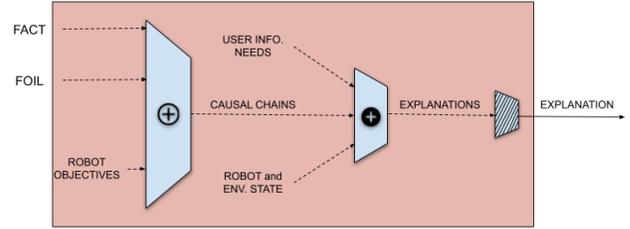

Figure 2: The explanation generation module. It receives the fact and foil as inputs and combines them with both robot's and environment's states to find causality chains (first sub-step). Those more related to the *explainer*'s objectives and *explainee*'s information needs are selected (second sub-step). Finally, filtering is used to select one explanation.

**Setting in Context** The second step aims to select a subset of explanations from the set produced in the previous step. These are chosen depending on both the *explainer*'s objective and *explainee*'s information need. A request for explanations is always due to a gap between the rationale behind the *explainer*'s action and the *explainee*'s understanding of such rationale. The *explainer* can figure out what this gap refers to by knowing, or inferring the *explainee*'s information needs. This requires that the robot would be able to build a reliable Theory of Mind (ToM) of the *explainee*: this feature could speed up the potentiality of the framework. Such ToM could simply consist of information about what the *explainee* perceives and knows: at least what the robot already explained to them. This aspect enables the *explainer* to choose, among all possible explanations, the ones that better fit with the *explainee*'s needs.

Devin and Alami (2016) used robots' ToMs in shared autonomy contexts to reduce unnecessary communication and produce a less intrusive robotic behaviour. However, more importantly for us, they theorised a possible use of robots' ToMs to estimate the lack of information of the robot or understand unexpected users' behaviours. In our opinion, this is not just customisation but a new and precise way to put the user at the centre of the explanation interaction.

The idea is similar to the Model Reconciliation presented by Chakraborti et al. (2017), where the two agents resolve the discrepancies between their internal models to find the most appropriate explanation. A similar approach has been used by Tabrez et al. (2019) for users' task understanding to detect incomplete or incorrect beliefs about the function-ing of the robot. Moreover, to select what information to give, above multiple choices, a Situation Awareness-based strategy has already been proposed (Sanneman and Shah 2020). However, in our opinion, a Shared Perception-based approach could turn out to be more appropriate (Kuhl 1998).

In addition to the users' information needs, this sub-step uses the robot's and environment's states to relate the explanation to the current context. In fact, a change in the environment's state is also due to humans actions, especially in collaborative contexts. Thus, considering the state of the environment means to consider also the effects of humans actions. Moreover, considering the robot state could be useful to relate its action to some particularly advantageous internal state.

**The Explanation Selection**   If more than one explanation fits the context, final filtering is necessary. This filtering should be based on heuristics taken from cognitive science: *e.g.*, to select recent causes (Miller and Gunasegaram 1990), or to prefer more controllable causes above non-intentional ones (Girotto et al. 1991). This way, also the level of detail is chosen.

### The Explanation Providing Module

Once the system has a good explanation, all that remains is to communicate it. The objective of this module is to choose *how* to provide the chosen explanation: it defines the explanation's shape and the communication modality. For what regards the latter, it depends on both the robot's communication skills and context: *e.g.*, the robot could provide for a vocal explanation or a graphical one via a tablet.

Also, the shape of the explanation is related to the robot's communication skills: *e.g.*, if the robot can speak, it has to choose the shape of the sentence. Once that the *explainer* has given the explanation, the *explainee* needs to interpret it to confer intentionality (ToM) to the robot's action. This step is key because, by putting together their ToMs, they can build a mutual understanding based on a common ground. An example of this attempt at alignment can be found in (Tabrez et al. 2019), where authors made the robot explain its reward function to give insights about its actions' rationale concerning the task.

### Example Scenario

Let us imagine a robot companion for elderly people that assists its owners in their everyday tasks. One of the robot's AI systems allows it to do the laundry. Hence, the robot decides to clean delicate clothes and starts collecting and bring them to the washing machine. In the meanwhile, the owner realises what the robot is doing but he can not understand why it left some clothes in the laundry hamper. Thus, he asks the robot why it did so. The robot gives the following explanation: "I did not take all the clothes because I am cleaning the delicates".

This simple scenario can be easily dropped into the context of our framework. Firstly, the XAI system receives the question "Why did not you take all clothes from the laundry hamper?" and produces the fact "the robot did not take some clothes" and the (implicit) foil "the robot took all clothes".

Subsequently, both the fact and foil are passed to the explanation generation module. Then, it produces causality chains but a good explanation is reached once inferred the gap between the robot action's rationale and the human understanding of the situation. In fact, the user does not know that the robot decided to clean only delicate clothes. In this case, no further filtering is needed. Finally, the robot chooses to verbalise its explanation and to structure it repeating the fact followed by "because" and then the explanation.

### Conclusions

In this work, we propose a general framework for user-centred explainable artificial intelligence (XAI) in human-robot interaction (HRI). The framework emphasises the social-interactive aspect of XAI by focusing on the communication between the agents involved. Also, our framework puts the person at the centre by exploiting well-known mechanisms taken from cognitive and social sciences. In particular, we defined it as user-centred because it considers the changes in the environment due to users action, their information need and human-inspired heuristics in the explanation selection. We focused on *post-hoc* explanations, which are those asked after the AI system's action. Theoretically, the framework can support any AI technique. In particular, the module that should be customised more for this purpose is the Explanation Generation one: the core of the entire framework. However, we think that our framework suits more with AI systems that admit interactions between *explainer* and *explainee* than with systems it is difficult to interact with. Concluding, we think that our work helps in building a bridge between XAI and HRI that is destined to be walked more and more in the years to come.

### Future Works

In addition to the implementation of the framework, from this work two possible research lines spring. The first one regards which explanation types (or strategies) better fit with HRI social contexts, as well as the role of robots' social behaviour in XAI. While the second one regards the solution to one of the most challenging issues that our framework arises: the users' information needs. As pointer out above, we think that this problem could be interestingly solved by using shared perception mechanisms. Therefore, we plan to follow these paths in the coming years.

### References

Adadi, A.; and Berrada, M. 2018. Peeking Inside the Black-Box: A Survey on Explainable Artificial Intelligence (XAI). *IEEE Access* 6: 52138–52160.

Arnold, T.; Kasenberg, D.; and Scheutz, M. 2021. Explaining in Time: Meeting Interactive Standards of Explanation for Robotic Systems. *ACM Transactions on Human-Robot Interaction* 10(3).

Chakraborti, T.; Sreedharan, S.; Zhang, Y.; and Kambhampati, S. 2017. Plan Explanations as Model Reconciliation: Moving Beyond Explanation as Soliloquy. In *Proceedings of the Twenty-Sixth International Joint Conference on Artificial Intelligence, IJCAI-17*, 156–163.

Chin-Parker, S.; and Bradner, A. 2010. Background shifts affect explanatory style: How a pragmatic theory of explanation accounts for background effects in the generation of explanations. *Cognitive processing* 11(3): 227–249.

Ciatto, G.; Schumacher, M. I.; Omicini, A.; and Calvaresi, D. 2020. Agent-Based Explanations in AI: Towards an Abstract Framework. In *Explainable, Transparent Autonomous Agents and Multi-Agent Systems*, 3–20. Springer International Publishing.

De Graaf, M. M.; and Malle, B. F. 2017. How people explain action (and autonomous intelligent systems should too). In *2017 AAAI Fall Symposium Series*.

Dennett, D. C. 1989. *The intentional stance*. MIT press.

Devin, S.; and Alami, R. 2016. An implemented theory of mind to improve human-robot shared plans execution. In *2016 11th ACM/IEEE International Conference on Human-Robot Interaction (HRI)*, 319–326. IEEE.

Ehsan, U.; Liao, Q. V.; Muller, M.; Riedl, M. O.; and Weisz, J. D. 2021. *Expanding Explainability: Towards Social Transparency in AI Systems*. Association for Computing Machinery.

Fair, D. 1979. Causation and the Flow of Energy. *Erkenntnis* 14(3): 219–250.

Girotto, V.; Legrenzi, P.; and Rizzo, A. 1991. Event controllability in counterfactual thinking. *Acta Psychologica* 78(1-3): 111–133.

Gunning, D.; and Aha, D. 2019. DARPA's explainable artificial intelligence (XAI) program. *AI Magazine* 40(2): 44–58.

Halpern, J. Y.; and Pearl, J. 2005. Causes and Explanations: A Structural-Model Approach. Part I: Causes. *The British Journal for the Philosophy of Science* 56(4): 843–887.

Han, Z.; Giger, D.; Allspaw, J.; Lee, M. S.; Admoni, H.; and Yanco, H. A. 2021. Building the Foundation of Robot Explanation Generation Using Behavior Trees. *ACM Transactions on Human-Robot Interaction* 10(3).

Heider, F. 2013. *The psychology of interpersonal relations*. Psychology Press.

Hellström, T.; and Bensch, S. 2018. Understandable robots - What, Why, and How. *Paladyn, Journal of Behavioral Robotics* 9(1): 110–123.

Hilton, D. J. 1990. Conversational processes and causal explanation. *Psychological Bulletin* 107(1): 65.

Hilton, D. J.; McClure, J. J.; and Slugoski, B. R. 2005. The course of events: counterfactuals, causal sequences, and explanation. Routledge.

Hume, D.; et al. 2000. *An enquiry concerning human understanding: A critical edition*, volume 3. Oxford University Press.

Kotsiantis, S. B.; Zaharakis, I. D.; and Pintelas, P. E. 2006. Machine learning: a review of classification and combining techniques. *Artificial Intelligence Review* 26(3): 159–190.

Kuhl, P. K. 1998. Language, culture and intersubjectivity: The creation of shared perception. .

Lewis, D. K. 1986. Causal explanation .

Lipton, P. 1990. Contrastive Explanation. *Royal Institute of Philosophy Supplement* 27: 247–266.

Lipton, Z. C. 2016. The mythos of model interpretability. CoRR abs/1606.03490 (2016). *arXiv preprint arXiv:1606.03490* .

Lombrozo, T. 2006. The structure and function of explanations. *Trends in Cognitive Sciences* 10(10): 464–470. ISSN 1364-6613.

Menzies, P.; and Price, H. 1993. Causation as a Secondary Quality. *The British Journal for the Philosophy of Science* 44(2): 187–203.

Miller, D. T.; and Gunasegaram, S. 1990. Temporal order and the perceived mutability of events: Implications for blame assignment. *Journal of personality and social psychology* 59(6): 1111.

Miller, T. 2019. Explanation in artificial intelligence: Insights from the social sciences. *Artificial Intelligence* 267: 1–38. ISSN 0004-3702.

Miller, T.; Howe, P.; and Sonenberg, L. 2017. Explainable AI: Beware of inmates running the asylum or: How I learnt to stop worrying and love the social and behavioural sciences. *arXiv preprint arXiv:1712.00547* .

Mohseni, S.; Zarei, N.; and Ragan, E. D. 2018. A multidisciplinary survey and framework for design and evaluation of explainable AI systems. *arXiv preprint arXiv:1811.11839* .

Montavon, G.; Samek, W.; and Müller, K.-R. 2018. Methods for interpreting and understanding deep neural networks. *Digital Signal Processing* 73: 1–15. ISSN 1051-2004.

Sanneman, L.; and Shah, J. A. 2020. A Situation Awareness-Based Framework for Design and Evaluation of Explainable AI. In *Explainable, Transparent Autonomous Agents and Multi-Agent Systems*, 94–110. Cham: Springer International Publishing.

Sokol, K.; and Flach, P. 2020. One explanation does not fit all. *KI-Künstliche Intelligenz* 34(2): 235–250.

Tabrez, A.; Agrawal, S.; and Hayes, B. 2019. Explanation-based reward coaching to improve human performance via reinforcement learning. In *2019 14th ACM/IEEE International Conference on Human-Robot Interaction (HRI)*, 249–257. IEEE.

Thellman, S.; and Ziemke, T. 2021. The Perceptual Belief Problem: Why Explainability Is a Tough Challenge in Social Robotics. *ACM Transactions on Human-Robot Interaction* 10(3).

Vilone, G.; and Longo, L. 2020. Explainable artificial intelligence: a systematic review. *arXiv preprint arXiv:2006.00093* .

Wallkötter, S.; Tulli, S.; Castellano, G.; Paiva, A.; and Chetouani, M. 2021. Explainable Embodied Agents Through Social Cues: A Review. *ACM Transactions on Human-Robot Interaction* 10(3).